 \documentclass[pmlr,twocolumn,10pt]{jmlr} 

\usepackage{booktabs}
\usepackage[load-configurations=version-1]{siunitx} 

\theorembodyfont{\upshape}
\theoremheaderfont{\scshape}
\theorempostheader{:}
\theoremsep{\newline}

\jmlrvolume{LEAVE UNSET}
\jmlryear{2021}
\jmlrsubmitted{LEAVE UNSET}
\jmlrpublished{LEAVE UNSET}
\jmlrworkshop{Machine Learning for Health (ML4H) 2021} 

\usepackage[utf8]{inputenc} 
\usepackage[T1]{fontenc}    
\usepackage{hyperref}       
\usepackage{url}            
\usepackage{booktabs}       
\usepackage{amsfonts}       
\usepackage{nicefrac}       
\usepackage{microtype}      
\usepackage{graphicx}
\usepackage{natbib}
\usepackage{doi}
\usepackage{caption}
\usepackage{subcaption}
\usepackage{natbib}

\usepackage{hyperref}       
\usepackage{url}            
\usepackage{amsfonts}       
\usepackage{nicefrac}       
\usepackage{xcolor}         
\usepackage{graphicx}
\usepackage{booktabs}
\usepackage{times}
\usepackage{amsfonts}
\usepackage{mathtools, enumitem}
\usepackage{latexsym}
\usepackage{relsize}
\usepackage{mathrsfs, float}
\usepackage{textcomp}
\usepackage{hyperref}
\usepackage{xcolor}
\usepackage{float}
\usepackage{wrapfig}
\usepackage{dsfont}
\usepackage{caption}
\usepackage{subcaption}
\usepackage{enumitem}

\definecolor{PrincetonOrange}{HTML}{ff8f00}

\newcommand{\defeq}{\stackrel{\text{def}}{=}}

\newcommand{\D}{\mathcal{D}}

\def\x{\mathbf{x}}

\newcommand{\ignore}[1]{}
\newcommand{\email}[1]{\texttt{#1}}

\DeclareMathAlphabet{\mathbfsf}{\encodingdefault}{\sfdefault}{bx}{n}

\newcommand{\E}{\mathbb{E}}

\newcommand{\eps}{\varepsilon}

\let\oldtfrac\tfrac
\renewcommand{\tfrac}[2]{\smash{\oldtfrac{#1}{#2}}}

\let\nablaold\nabla
\renewcommand{\nabla}{\nablaold\mkern-2.5mu}

 \title{Machine Learning for Mechanical Ventilation Control}

\author{
  \Name{Daniel Suo}\thanks{Google LLC}\thanks{Princeton University},
  \Name{Naman Agarwal}\footnotemark[1],
  \Name{Wenhan Xia}\footnotemark[1]\footnotemark[2],
  \Name{Xinyi Chen}\footnotemark[1]\footnotemark[2],
  \Name{Udaya Ghai}\footnotemark[1]\footnotemark[2],
  \Name{Alexander Yu}\footnotemark[1],
  \Name{Paula Gradu}\footnotemark[1],
  \Name{Karan Singh}\footnotemark[1]\footnotemark[2],
  \Name{Cyril Zhang}\footnotemark[1]\footnotemark[2],
  \Name{Edgar Minasyan}\footnotemark[1]\footnotemark[2], 
  \Name{Julienne LaChance}\footnotemark[2], 
  \Name{Tom Zajdel}\footnotemark[2], 
  \Name{Manuel Schottdorf}\footnotemark[2], 
  \Name{Daniel Cohen}\footnotemark[2], 
  \Name{Elad Hazan}\footnotemark[1]\footnotemark[2] 
}

\begin{document}

\maketitle

\begin{abstract}
Mechanical ventilation is one of the most widely used therapies in the ICU. However, despite broad application from anaesthesia to COVID-related life support, many injurious challenges remain.

We frame these as a control problem: ventilators must let air in and out of the patient's lungs according to a prescribed trajectory of airway pressure. Industry-standard controllers, based on the PID method, are neither optimal nor robust.

Our data-driven approach learns to control an invasive ventilator by training on a simulator itself trained on data collected from the ventilator. This method outperforms popular reinforcement learning algorithms and even controls the physical ventilator more accurately and robustly than PID.

These results underscore how effective data-driven methodologies can be for invasive ventilation and suggest that more general forms of ventilation (e.g., non-invasive, adaptive) may also be amenable.
\end{abstract}

\section{Introduction}

Mechanical ventilation is a widely used treatment with applications spanning anaesthesia \citep{coppola2014protective}, neonatal intensive care \citep{van2019modes}, and life support during the current COVID-19 pandemic \citep{meng2020intubation, wunsch2020mechanical, mohlenkamp2020ventilation}. Despite its use in ICUs for decades, mechanical ventilation can still lead to ventilator-induced lung injury (VILI) for patients \citep{vili}.

\begin{figure}[!h]
\centering

\begin{subfigure}{.49\textwidth}
     \centering
     \includegraphics[width=\linewidth]{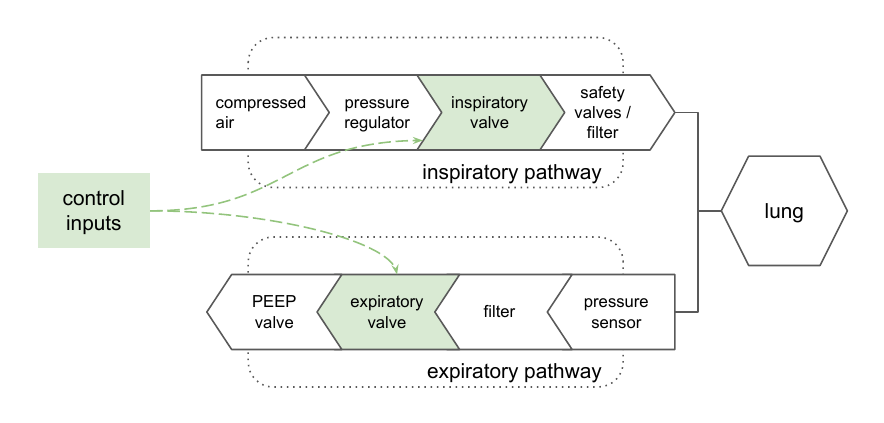}
     \subcaption{A simplified respiratory circuit showing the airflow through the inspiratory pathway, into and out of the lung, and out the expiratory pathway. We shade the components that our algorithms can control in green.}
     \label{fig:circuit}
 \end{subfigure}\hfill%
 \begin{subfigure}{.49\textwidth}
     \centering
     \includegraphics[width=\linewidth]{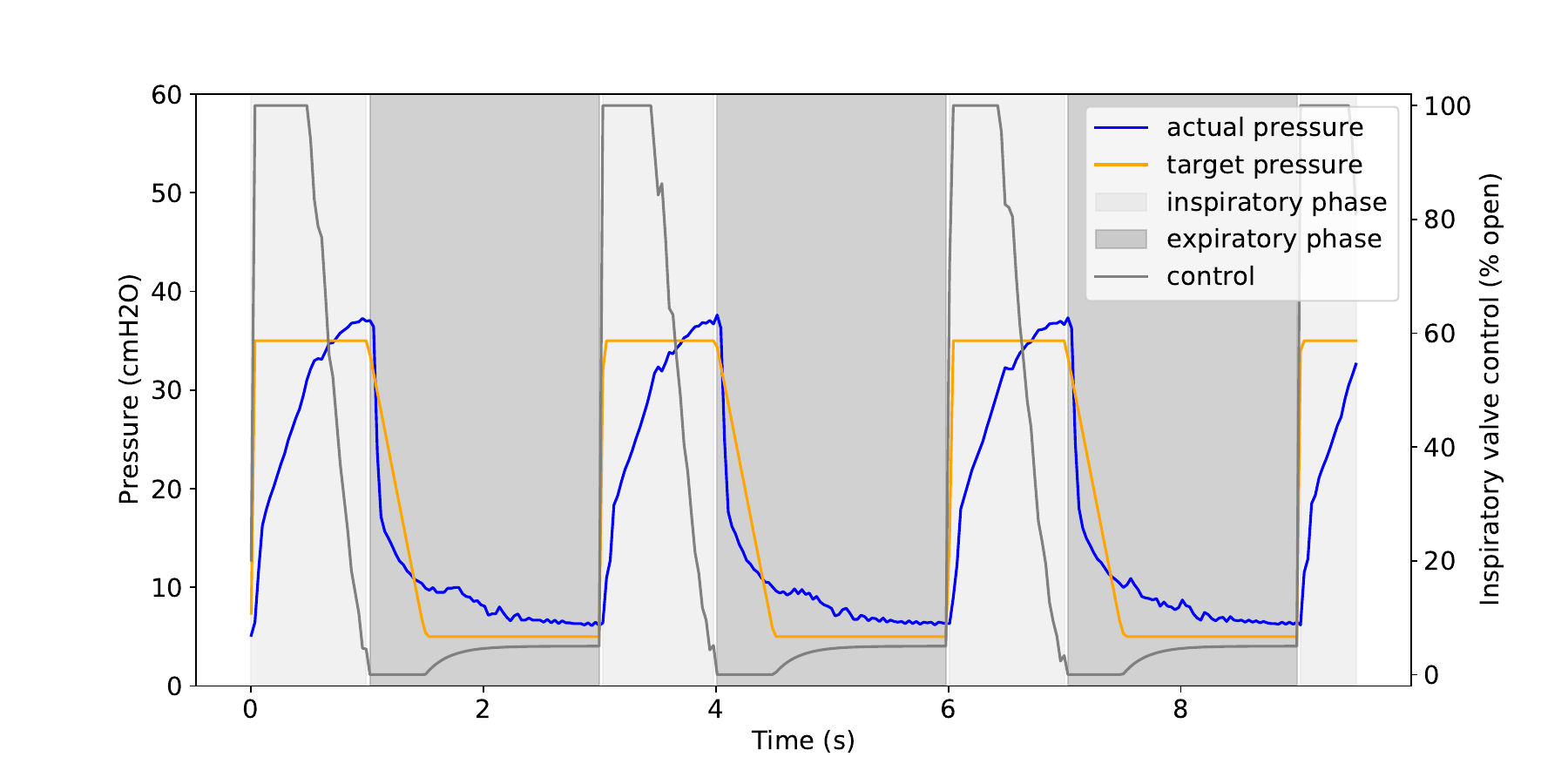}
     \subcaption{An example run of three breaths where PID (dark gray line) controls lung pressure (blue line) according to a prescribed target waveform (orange line).}
     \label{fig:tracking}
\end{subfigure}
\end{figure}

We focus on pressure-controlled invasive ventilation (PCV) \citep{rittayamai2015pressure}. In this setting, an algorithm controls two valves (see Figure \ref{fig:circuit}) that let air in and out of a fully-sedated patient's lung according to a target waveform of lung pressure (see Figure \ref{fig:tracking}). We consider the control task only on ISO-standard \citep{ISO68844} artificial lungs.

Despite its importance, ventilator control has remained largely unchanged for years, relying on PID \citep{pid2} controllers and similar variants to track patient state according to a prescribed target waveform. A ventilator controller must adapt quickly and reliably across the spectrum of clinical conditions, which are only indirectly observable given a single measurement of pressure.

A model that is highly expressive may learn the dynamics of the underlying systems more precisely and thus adapt faster to the patient's condition. However, such models usually require a large amount of data to train, which can take prohibitively long to collect by purely running the ventilator. We opt instead to learn a simulator to generate artificial data, though learning such a simulator for a partially observed non-linear system is itself a difficult problem. As part of our contributions, we present better-performing, more robust results and present resources for future researchers. 

\section{Scientific background}
\label{sec:background}

\paragraph{Control of dynamical systems} We begin with some formalisms of the control problem. A partially-observable discrete-time dynamical system is given by the following equation: 
$ x_{t+1} = f(x_t, u_t), o_{t+1} = g(x_{t+1})$,
where  $x_t$ is the underlying state of the dynamical system, $o_t$ is the observation of the state available to the controller, $u_t$ is the control input and $f,g$ are the transition function and observation functions respectively. Given a dynamical system, the control problem is to minimize the sum of cost functions over a long-term horizon:
$\min_{u_{1:T} }  \sum_{t=1}^T c_t(x_t, u_t) \quad \text{s.t.}\;\; x_{t+1} = f_t(x_t, u_t)$.

A ubiquitous technique for the control of dynamical systems is the use of linear error-feedback controllers, i.e. policies that choose a control based on a linear function of the current and past errors vs. a target state. That is,
$ u_{t+1} = \sum_{i=0}^k \alpha_i \eps_{t-i}$,
where $\eps_t = x_t - {x}^\star_t$ (or $\eps_t = o_t - {o}^\star_t$ if the system is partially-observable) is the deviation from the target state at time $t$, and $k$ represents the history length of the controller. PID applies a linear control with \emph{proportional}, \emph{integral}, and \emph{differential} coefficients,
$ u_t = \alpha \eps_{t} + \beta \sum_{i=0}^k \eps_{t-i} + \gamma (\eps_{t} - \eps_{t-1}) $.
This special class of linear error-feedback controllers, motivated by physical laws, is a simple, efficient and widely used technique  \citep{PID1}. It is currently the industry standard for (open-source) ventilator control.  

\paragraph{The physics of ventilation}
The goal of ventilator control is to regulate the pressure sensor measurements to follow a target waveform $p_t^{\star}$ via controlling the air-flow into the system which forms the control input $u_t$. As a dynamical system, we can denote the underlying state of the ventilator-patient system as $x_t$ evolving as $x_{t+1} = f(x_t, u_t),$ for an unknown $f$ and the pressure sensor measurement $p_t$ is the observation available to us. The cost function can be defined to be a measure of the deviation from the target; e.g. the absolute deviation $c_t(p_t, u_t) = |p_t - p_t^{\star}|$. The objective is to design a controller that minimizes the total cost over $T$ time steps. 

\paragraph{Challenges and benefits of a model-based approach} Physics-based dynamics models (\citep{nadeem_2021}) rely on idealized physical behaviors, lagged and partial observations, and underspecification of how the true system may vary. As a result, we adopt a learned model-based approach due to its sample-efficiency and reusability. A reliable simulator enables much cheaper and faster data collection for training a controller, and allows us to incorporate multitask objectives and domain randomization (e.g. different waveforms, or even different patients). An additional goal is to make the simulator \emph{differentiable}, enabling direct gradient-based policy optimization through the system's dynamics (rather than stochastic estimates thereof).

We show that in this partially-observed (but single-input single-output) system, we can query a reasonable amount of training data in real time from the test lung, and use it offline to learn a differentiable simulator of its dynamics (\emph{``real2sim''}). Then, we complete the pipeline by leveraging interactive access to this simulator to train a controller (\emph{``sim2real''}). We demonstrate that this pipeline is sufficiently robust that the learned controllers can outperform PID controllers tuned directly on the test lung.

\section{Experimental Setup}
\label{sec:hardware}

\paragraph{Test lung and ventilator}
To develop simulators and control algorithms, we run mechanical ventilation tasks on a physical test lung \citep{ingmar_medical_2020} using the open-source ventilator designed by Princeton University's People's Ventilator Project (PVP) \citep{pvp2020}.

\paragraph{Abstraction of the simulation task} We treat the mechanical ventilation task as episodic by separating each inspiratory phase from the breath timeseries and treating those as individual episodes. This approach reflects both physical and medical realities. Mechanically ventilated breaths are by their nature highly regular and feature long expiratory phases  that end with the ventilator-lung system close to its initial state, thereby justifying the episodic nature. Further, the inspiratory phase is indeed the most relevant to clinical treatment and the harder regime to control with prevalent problems of under- or over-shooting the target pressure and ringing.
\begin{figure}[!h]
    \centering
    \begin{subfigure}{0.49\textwidth}
    \centering
    \includegraphics[width=0.6\linewidth]{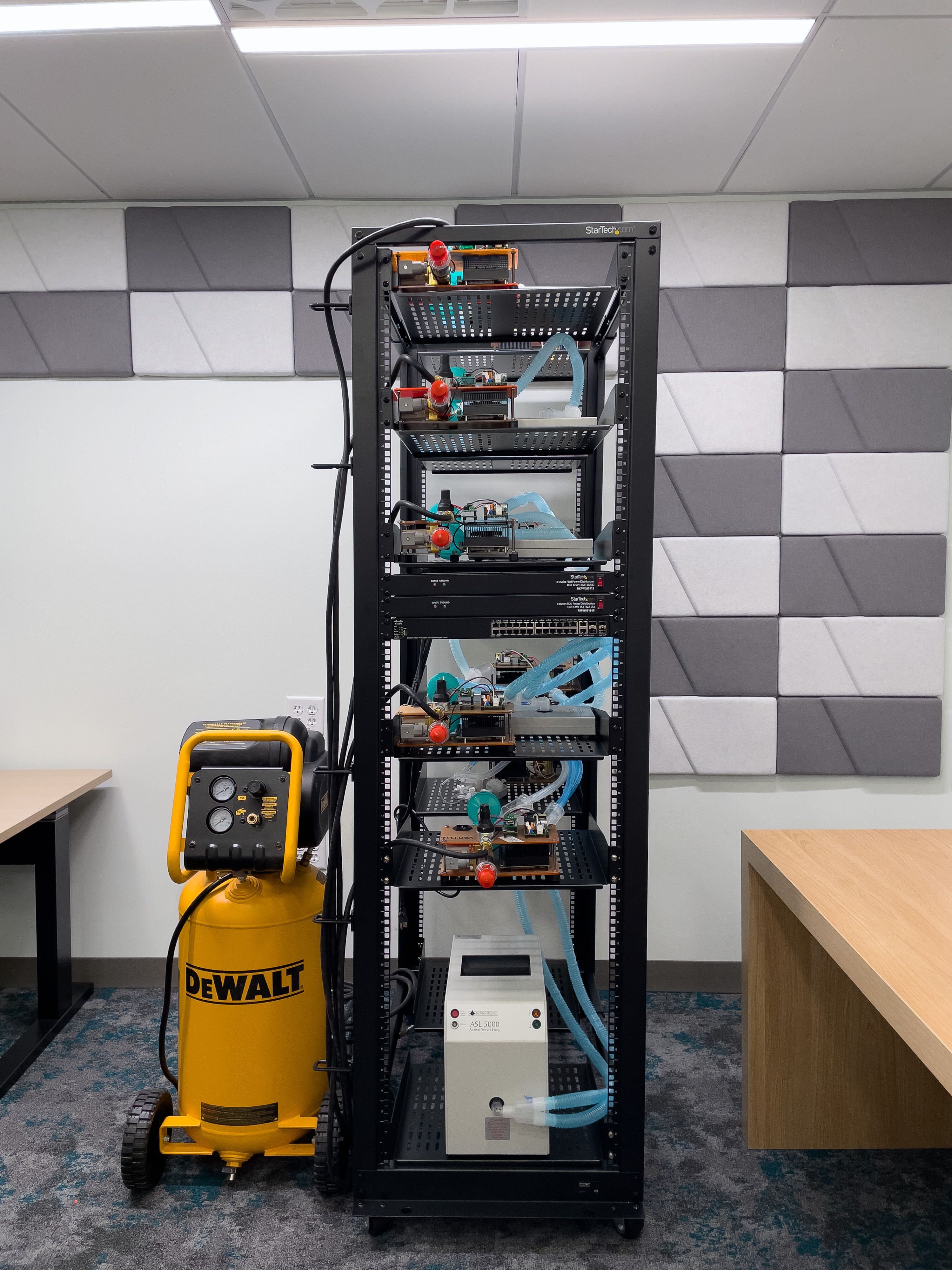}
    \caption{The ventilator cluster we constructed to run our experiments, featuring 10 ventilators, 4 air compressors, and 2 control servers. Each ventilator is re-calibrated after each experimental run for consistency across ventilators and over time.}
    \label{fig:vent-farm}
    \end{subfigure}\hfill%
    \begin{subfigure}{0.49\textwidth}
\centering
\includegraphics[width=\linewidth]{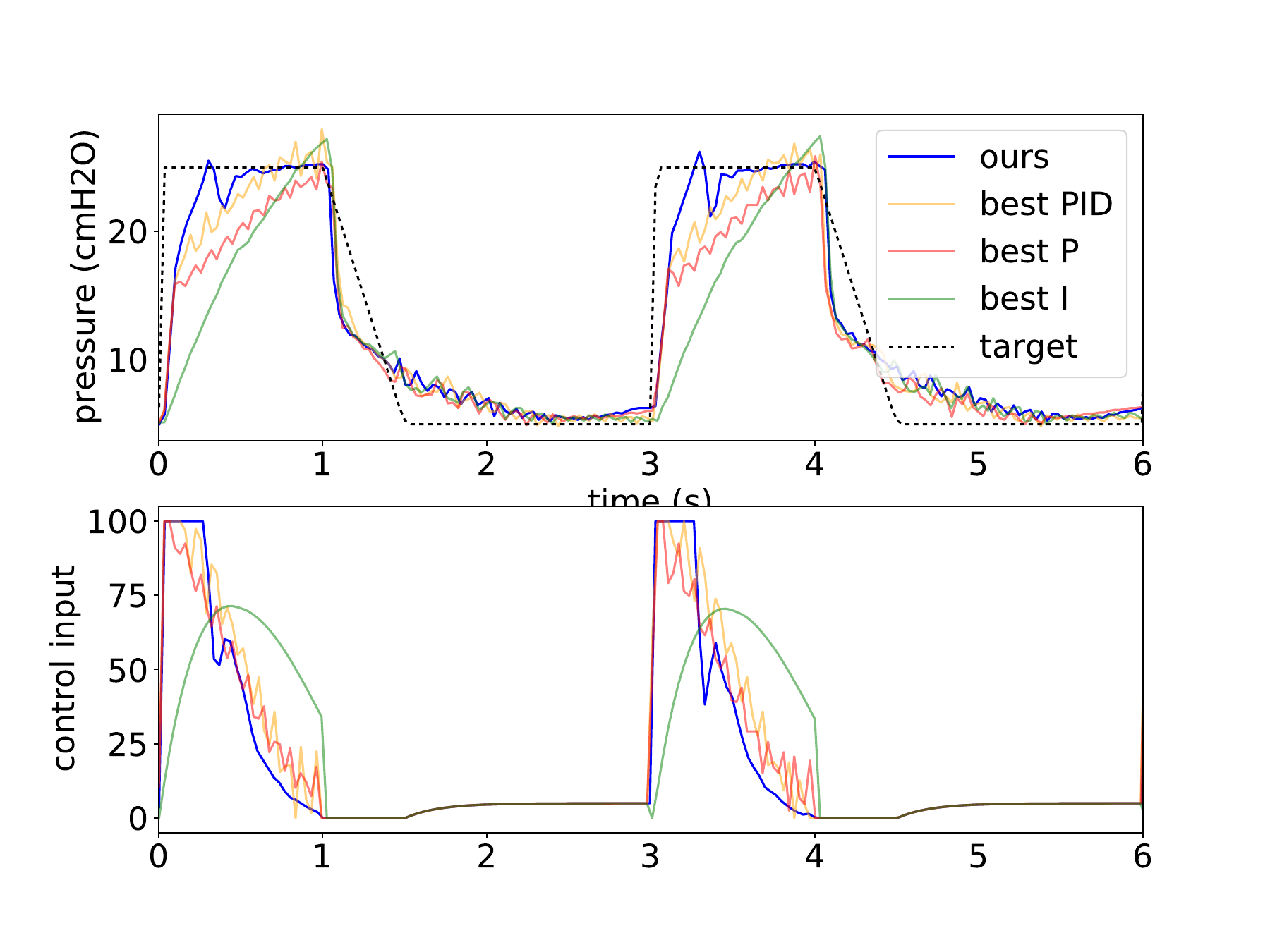}
\caption{As an example, we compare our method (learned controller on learned simulator) to the best P-only, I-only, and PID controllers relative to a target waveform (dotted line). Whereas our controller rises quickly and stays very near the target waveform, the other controllers take significantly longer to rise, overshoot, and, in the case of P-only and PID, ring the entire inspiratory phase.}\label{fig:traj}
\end{subfigure}
\end{figure}

\section{Learning a data-driven simulator} \label{sec:sim}

The evaluation for any simulator can only be performed using a {\bf black-box metric}, since we do not have explicit access to the system dynamics, and existing physics models are poor approximations to the empirical behavior. 
    
Let $\D$ be a distribution over sequences of controls denoted $\mathbf{u} = \{u_1,u_2,...,u_T\}$.

We define the {\bf open-loop distance} w.r.t. horizon $T$ and control sequence distribution $\D$ as
    $\|f_1 - f_2\|_{ol} \defeq \E_{\mathbf{u} \sim \D} \left[ \sum_{t=1}^T \| f_1(x_{t,1},u_t) - f_2(x_{t,2},u_t) \|  \right]$.


\paragraph{Data collection} balance physical safety against the need to explore the space of controls and resulting pressures by choosing a known good PID for a given lung setting and introducing random exploratory perturbations according to the following two policies:
\begin{enumerate}
    \item Boundary exploration: To the very beginning of the inhalation, add an additional control sampled uniformly from $(c^a_{\min}, c^a_{\max})$ and decrease this additive control linearly to zero over a time frame sampled randomly from $(t^a_{\min}, t^a_{\max})$. This policy capitalizes on the fact that the beginning of a breath is safer and exploration here is more valuable since it is the driving phase of inspiration. 
    \item Triangular exploration: sample a maximal additional control from a range $(c^b_{\min}, c^b_{\max})$ and an interval $(t^b_{\min}, t^b_{\max})$, within the inhalation. Start from $0$ additional control at time $t^b_{\min}$, increase the additional control linearly until $(t^b_{\min} + t^b_{\max}))/2$, and then decrease to $0$ linearly until $t^b_{\min}$. This policy helps explore the intrinsic delay in the system between a control and a resulting change in system state.
\end{enumerate}

For each breath during data collection, we choose policy $(a)$ with probability $p_a$ and policy $(b)$ with probability $(1-p_a)$. The ranges in $(a)$ and $(b)$ are lung-specific.

\paragraph{Training Task(s).} The simulator aims to learn the unknown dynamics of the inhalation phase. We approximate the state of the system (which is not observable to us) by the sequence of the past pressures and controls upto a history length of $H_c$ and $H_p$ respectively. The task of the simulator can now be distilled down to that of predicting the next pressure $p_{t+1}$, based on the past $H_c$ controls $u_{t},\ldots, u_{t-H_c}$ and $H_p$ pressures $p_{t},\ldots, p_{t-H_p}$. We define the training task by constructing a regression data set whose inputs come from contiguous overlapping sections of $H_p, H_c$ within the collected trajectories and the task is to predict the following pressure.

\section{Learning controllers from learned physics} \label{sec:control}
\begin{figure}[!h]
\centering
\begin{subfigure}{0.49\textwidth}
\centering 
\includegraphics[width=\linewidth]{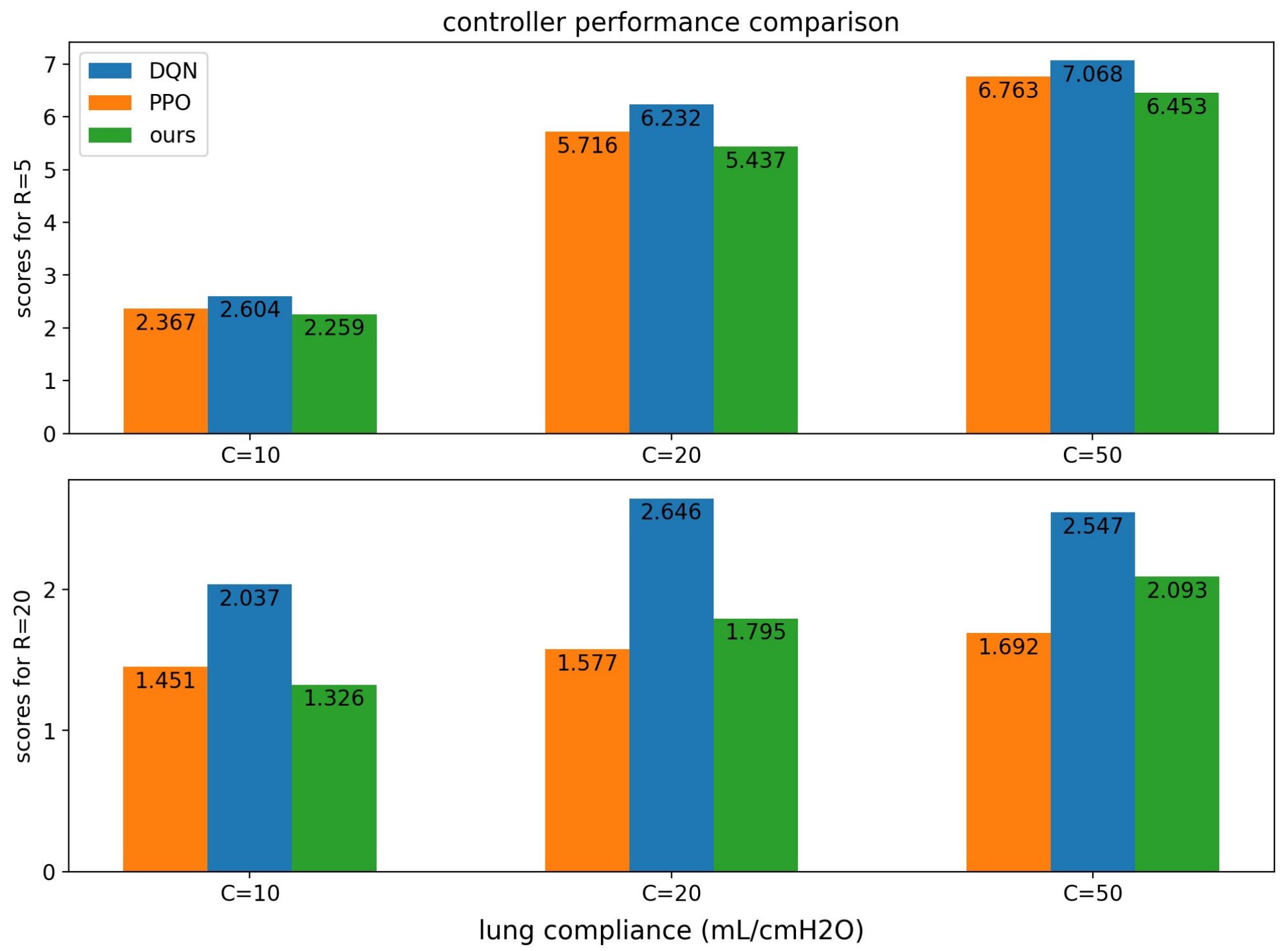}
\subcaption{Performance comparison of our controller with PPO/DQN. The score is calculated by average per-step L1 distance between target and achieved pressure.}\label{fig:baselines}
\end{subfigure}\hfill%
\begin{subfigure}{0.49\textwidth}
\centering 
\includegraphics[width=\linewidth]{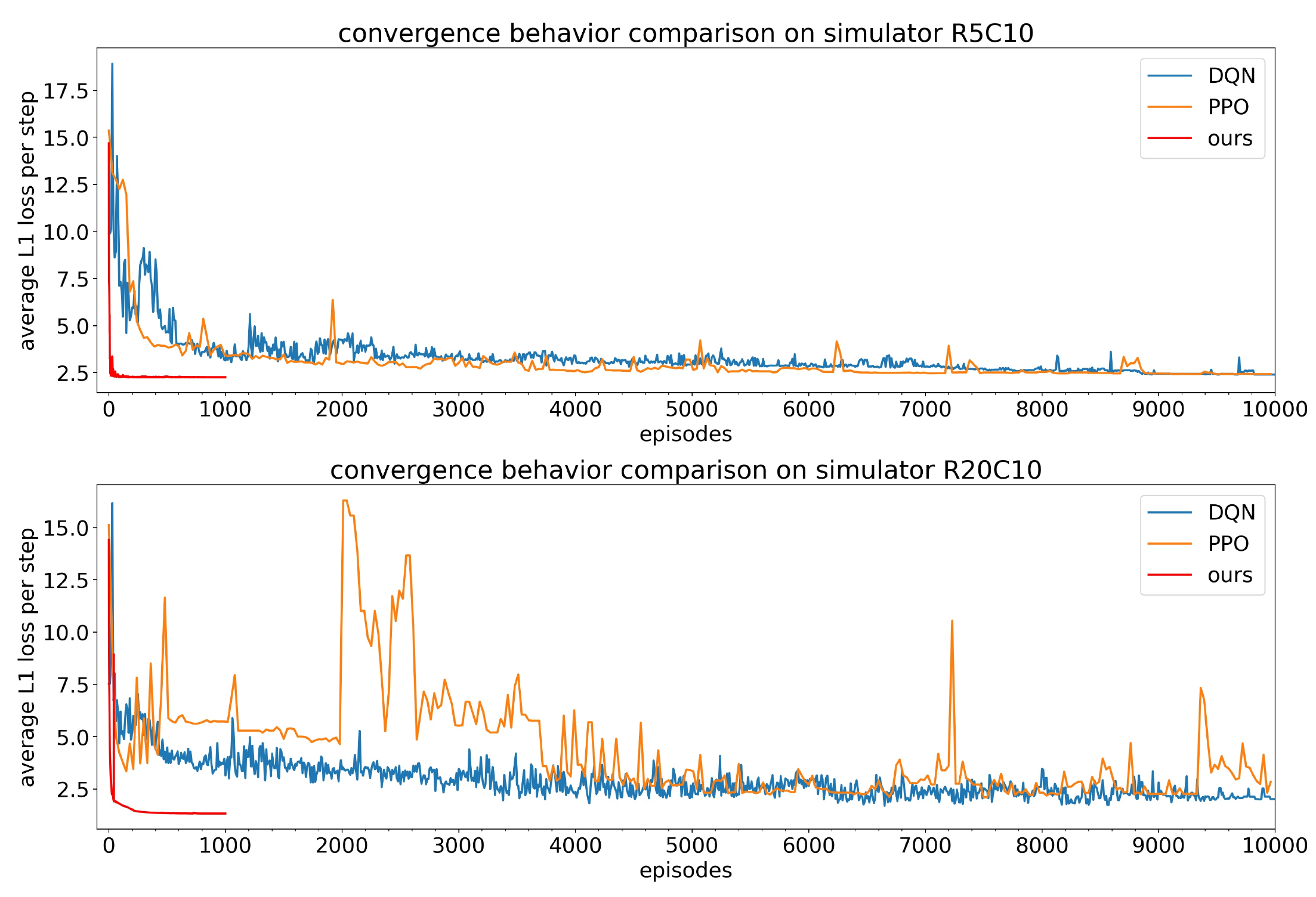}
\subcaption{Convergence behavior demonstration. Our methods converge using orders of magnitudes fewer samples, even when considering training data for our simulator.}\label{fig:samplecomplexity}
\end{subfigure}
\end{figure}

We focus on the following two tasks: (1) {\bf Performance:} improve performance for tracking desired waveform in ISO-specified benchmarks. Specifically, we minimize the combined $L_1$ deviation from the target inhalation behavior across all target pressures on the simulator corresponding to a single lung setting of interest. We also compare performance against several well-studied RL algorithms (Figure \ref{fig:baselines}). (2) {\bf Robustness:} improve performance using a {\bf single} trained controller. Specifically, we minimize the combined $L_1$ deviation from the target inhalation behavior across all target pressures \textit{and} across the simulators corresponding to \textit{several} lung settings of interest.

\paragraph{Controller architecture} 
Our controller is comprised of a PID baseline upon which we learn a deep network correction controlled with a regularization parameter $\lambda$. This \textit{residual} setup can be seen as a regularization against the gap between the simulator and the real dynamics. In particular this prevents the controller training from over-fitting on the simulator. We found this approach to be significantly better than the directly using the best (and perhaps over-fitted) controller on the simulator.

\paragraph{Experiments} To make comparisons between PID and our controllers on the physical system, we compute a score for each controller on a given test lung setting (e.g., $R=5, C=50$) by averaging the $L_1$ deviation from a target pressure waveform for all inspiratory phases, and then averaging these average $L_1$ errors over six waveforms specified in \citet{ISO68844}. We choose $L_1$ as an error metric so as not to over-penalize breaths that fall short of their target pressures and to avoid engineering a new metric. We determine the best performing PID controller for a given setting by running exhaustive grid searches over $P,I,D$ coefficients for each lung setting (details for both our score and the grid searches can be found in the Appendix).

\begin{figure}[!h]
\centering
\begin{subfigure}{0.49\textwidth}
\centering 
\includegraphics[width=\linewidth]{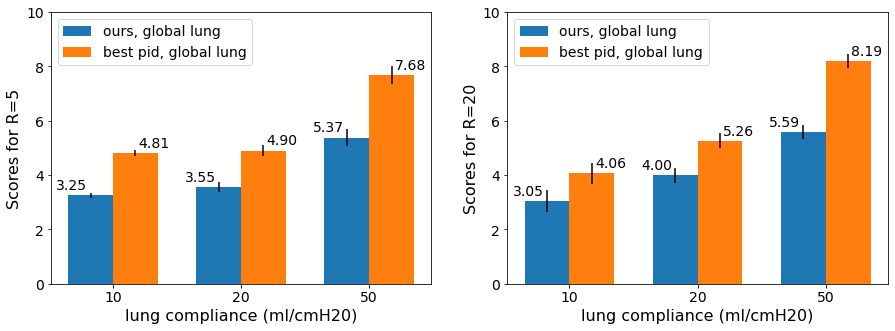}
\subcaption{We show that for each lung setting, the controller we trained on the simulator for that setting outperforms the best-performing PID controller found on the physical test lung.}\label{fig:global}
\end{subfigure}\hfill%
\begin{subfigure}{0.49\textwidth}
\centering 
\includegraphics[width=\linewidth]{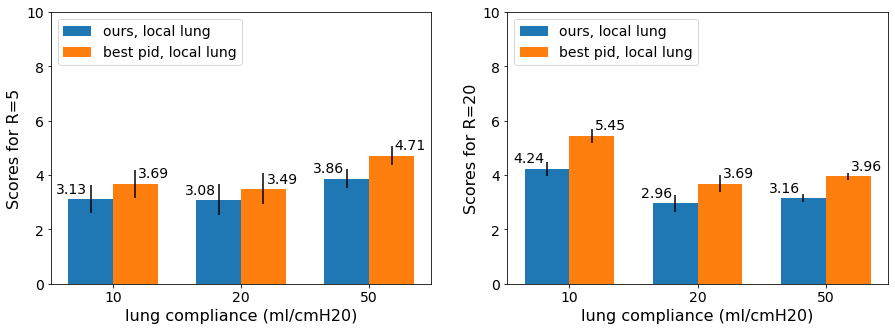}
\subcaption{The controller we trained on all six simulators outperforms the best PID found over the the same six settings on the physical test lung. Of note, our wins are proportionally greater when trained on all six settings whereas individual lung settings are more achievable by PID alone.}\label{fig:local}
\end{subfigure}
\end{figure}

\section{Conclusions and future work} \label{sec:discussion}

We have presented a machine learning approach to ventilator control, demonstrating the potential of end-to-end learned controllers by obtaining improvements over industry-standard baselines.

There remain a number of areas to explore, mostly motivated by medical need. The lung settings we examined are by no means representative of all lung characteristics (e.g., neonatal, child, non-sedated) and lung characteristics are not static over time; a patient may improve or worsen, or begin coughing. Ventilator costs also drive further research. As an example, inexpensive valves have less consistent behavior and longer reaction times, which exacerbate bad PID behavior (e.g., overshooting, ringing), yet are crucial to bringing down costs and expanding access. Learned controllers that adapt to these deficiencies may obviate the need for such trade-offs.

\bibliography{vent}

\end{document}